\documentclass[a4paper,conference]{IEEEtran}

\usepackage{graphicx, multirow}
\usepackage{amsmath}
\usepackage{hyperref}
\usepackage{subfigure}

\usepackage{xspace}
\makeatletter
\DeclareRobustCommand\onedot{\futurelet\@let@token\@onedot}
\def\@onedot{\ifx\@let@token.\else.\null\fi\xspace}

\def\eg{\emph{e.g}\onedot} 
\def\ie{\emph{i.e}\onedot} 
 
\def\etc{\emph{etc}\onedot} 
 
\def\etal{\emph{et al}\onedot}
\makeatother

\begin{document}
\title{Human Pose Estimation from RGB Input Using Synthetic Training Data}
\date{}

\author{Oscar Danielsson and Omid Aghazadeh\\
School of Computer Science and Communication\\
KTH, Stockholm, Sweden\\
{\tt\small \{osda02, omida\}@kth.se}
}
\maketitle

%

\begin{abstract}
We address the problem of estimating the pose of humans using RGB image input. More specifically, we are using a random forest classifier to classify pixels into joint-based body part categories, much similar to the famous Kinect pose estimator \cite{Shotton:cvpr:11, Shotton:pami:12}. However, we are using pure RGB input, i.e. no depth. Since the random forest requires a large number of training examples, we are using computer graphics generated, synthetic training data. In addition, we assume that we have access to a large number of real images with bounding box labels, extracted for example by a pedestrian detector or a tracking system. We propose a new objective function for random forest training that uses the weakly labeled data from the target domain to encourage the learner to select features that generalize from the synthetic source domain to the real target domain. We demonstrate on a publicly available dataset \cite{Kazemi2011} that the proposed objective function yields a classifier that significantly outperforms a baseline classifier trained using the standard entropy objective \cite{Quinlan86}.
\end{abstract}
\section{Introduction and related work}
\label{sec:intro}
Human pose estimation is a well studied problem in computer vision. 
Current state of the art methods can be divided into two groups: 1) those that explicitly model relative body part locations \eg \cite{Yang:cvpr11}, and 2) data driven methods that do not explicitly model valid poses \eg \cite{Shotton:pami:12, Kazemi2011}. 
Motivated by the increasing availability of large datasets \cite{HalevyNP09}, we are interested in data driven approaches using non-parametric and scalable classifiers.
Particularly, random forests  have become popular for human pose estimation.
The main reasons for the success of random forests are: 1) scalable training costs, 2) computationally efficient testing, and 3) adaptive complexity controlled by the depth of the trees.

While the flexibility and non-parametric nature of random forests enables them to solve difficult classification problems, it also limits their applicability to cases where one has access to training data from exactly the same distribution as the test data. 
In the case of football, this implies a need for labelled training data from the \emph{same arena}, with the \emph{same teams} playing at the \emph{same time of day} with the \emph{same advertisement signs} in the background. 
This makes such a system expensive and impractical. 
The method described in this paper provides practitioners with one tool to extend the applicability of random forest classifiers.

\begin{figure} [t]
  	\centering
	\subfigure[Current tree node  $n$]{
		\label{fig:intro:treenode}\includegraphics[width=0.3\linewidth]{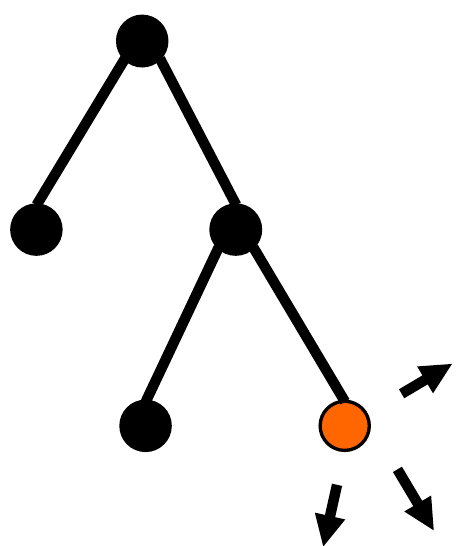}
	}~
	\subfigure[Spatial distribution of real training pixels at node $n$]{
		\label{fig:intro:real_spdist}\includegraphics[width=0.3\linewidth]{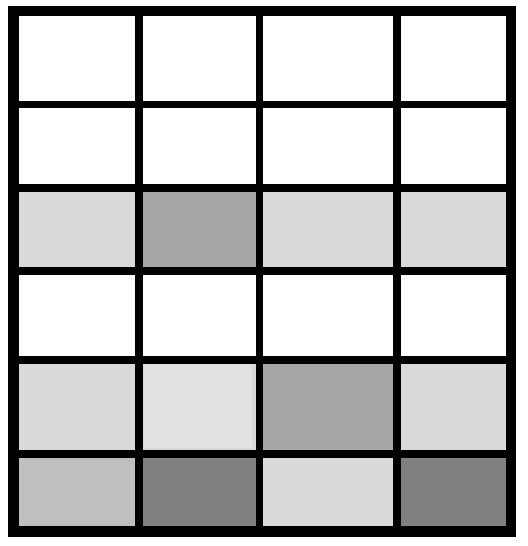} 
	}~
	\subfigure[Example real training image (unlabelled)]{
		\label{fig:intro:real_img}\includegraphics[width=0.3\linewidth]{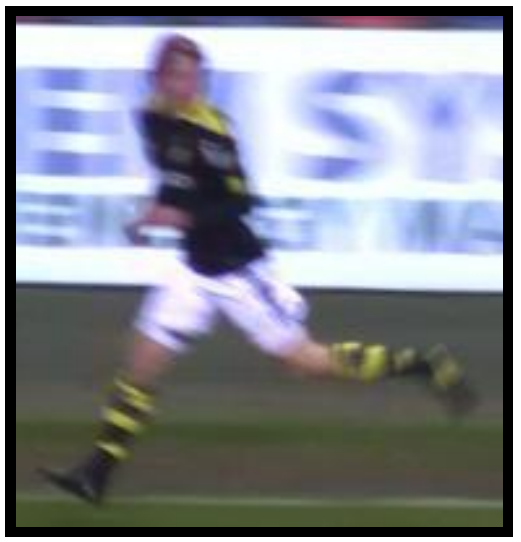}
	} \\
	\subfigure[Body part label distribution of synthetic training pixels at node $n$]{
		\label{fig:intro:syn_partdist}\includegraphics[width=0.3\linewidth]{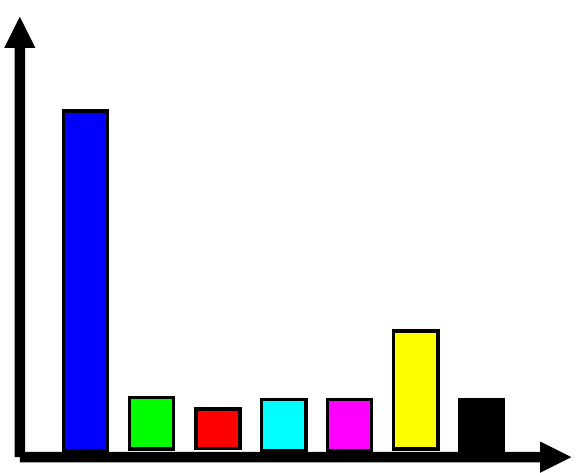} 
	}~
	\subfigure[Spatial distribution of synthetic training pixels at node $n$]{
		\label{fig:intro:syn_spdist}\includegraphics[width=0.3\linewidth]{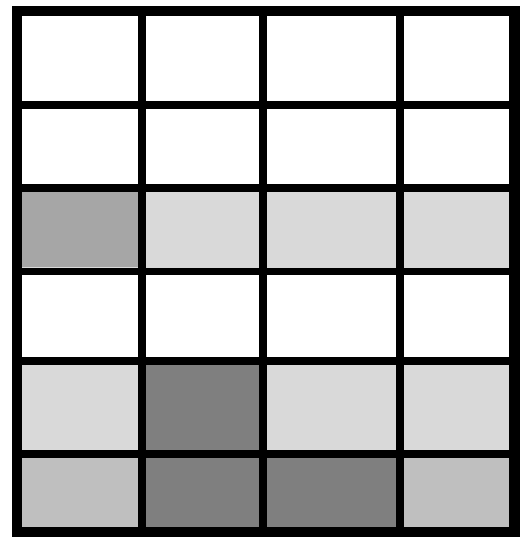} 
	}~
	\subfigure[Example synthetic training image (labelled)]{
		\label{fig:intro:syn_img}\includegraphics[width=0.3\linewidth]{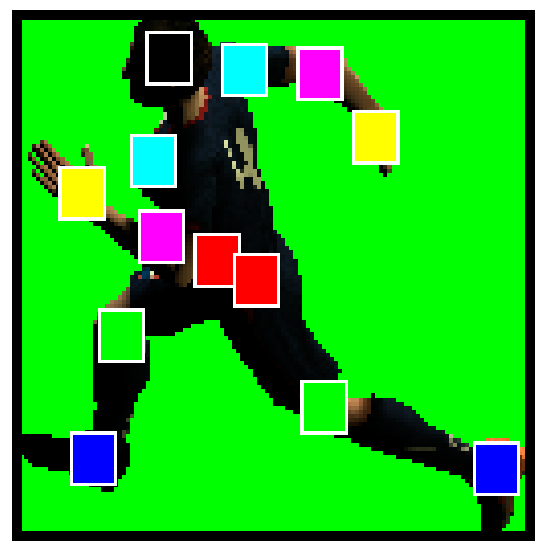}
	}\\
  \caption{
  Illustration of our approach. 
  We train a random forest pixel classifier using strongly labelled synthetic training images \ref{fig:intro:syn_img} and weakly labelled real images \ref{fig:intro:real_img}. 
  The synthetic training images have bounding box annotations and body part labels, whereas the real training images have only bounding box annotations. 
  We define the fitness $f(n)$ of a decision tree node $n$ \ref{fig:intro:treenode} as a linear combination of 1) the entropy of the label histogram of all labelled (synthetic) training pixels reaching the node \ref{fig:intro:syn_partdist}, and 2) the distance between the spatial distribution of real training pixels \ref{fig:intro:real_spdist} and the spatial distribution of the synthetic training pixels \ref{fig:intro:syn_spdist}.
  }
  \label{fig:intro_adaptation}
\end{figure}

In addition to requiring the training data and test data to be similar (\ie from the same distribution), the lack of generalization also requires the training data to span all the sources of variation in the test data.
There are usually many sources of variation involved in the imaging process \eg viewpoint, articulation, body shape, foreground texture, background texture, lighting conditions, \etc.
Consequently, the number of required samples for random forest classifiers to perform well is usually very large \eg millions of synthetic depth images were used to train the famous Kinect random forest classifier \cite{Shotton:cvpr:11}.

The most cost efficient way to acquire large annotated training sets is to use a synthetic model to sample arbitrarily many labelled training examples.
While generating realistic synthetic depth images at a low resolution is possible \cite{Shotton:cvpr:11}, the domain of synthesized RGB images is very different from that of the real images aquired by a camera, see figure \ref{fig:intro_adaptation} for an example.
Therefore, a random forest classifier trained on the synthetic domain cannot be expected to work well on the real domain.


The main contribution of this paper is to propose a way to encourage the random forest training procedure to select weak classifiers that ``behave similarly'' in source and target domains.
More specifically we modify the objective function for training a decision tree node such that the resulting weak classifier is encouraged to have consistent spatial activation pattern across domains.
This requires only a weak labelling of the samples from the target domain (real images) in form of a bounding box annotation. 
Often this can be extracted automatically by a pedestrian detector, tracking system or simple background subtraction.
This is illustrated in figure \ref{fig:intro_adaptation}.

The rest of the paper is structured as follows. 
We review and compare to related work in section \ref{sec:intro:related_work}.
In section \ref{sec:training} we describe our proposed objective function for random forest training and some implementation details of our system.
In section \ref{sec:experiments} we show that random forests trained with this objective function outperform random forests trained with the standard entropy objective. 
Finally we conclude in section \ref{sec:conclusion}.

\subsection{Related Work}
\label{sec:intro:related_work}
The problem of human pose estimation from images has received a lot of attention, see \cite{moeslund2011visual} for a good overview.
Notable uses of random forest classifiers for pose estimation include Kazemi and Sullivan \cite{Kazemi2011}, who applied random forests to pose estimation in RGB images. 
However, they used very similar data for training and testing, and consequently avoiding the domain adaptation problem. 
Shotton \etal did use training and test data from different sources, but used depth images instead of RGB to avoid the domain difference \cite{Shotton:cvpr:11}.
Pishchulin \etal proposed a sophisticated generative model to augment strongly labelled training data with synthetic RGB images of novel poses \cite{pishchulin12cvpr}.
They show a significant boost in performance using their data augmentation approach for both human detection and pose estimation tasks, using deformable part models and pictorial structures.
In contrast, all our labelled images are synthetic and we address domain shift explicitly in the objective function of random forest training.

The domain adaptation problem has been studied in various contexts in vision before. 
Kulis \etal proposed a method to learn asymmetric kernel transforms, that map data points from the source domain to the target domain, using labelled training data from both domains \cite{Kulis:cvpr:11}. 
Becker \etal use boosting to learn non-linear mappings from source domains to a shared feature space \cite{Becker_NIPS_2013}.
Yang \etal used an auxiliary classifier to adapt an existing classifier to a new domain \cite{Yang:acm:07}. 
Hoffman \etal used a linear transformation which in combination with a linear classifier performs domain adaptation \cite{DIAdaptation}.
The basic assumptions of these approaches are that the source feature spaces have known and finite dimensionality, and that the parameters of the transformation from source to target domain can be learned using a relatively small number of labelled training examples from the target domain. 
These assumptions are not well satisfied in our case, since the features are generated as a part of the training algorithm.

\section{Random forest training}
\label{sec:training}
We want to train a random forest to classify pixels into a number of body part classes. Each body part is localized around a joint and we have defined 8 body parts, including a background class. The classes are: `foot', `knee', `hip', `shoulder', `elbow', `hand', `head' and `background' (see figure \ref{fig:intro:syn_img}, which depicts all classes except the background). Our training data consists of a number of fully labelled synthetic images and a number of unlabelled real images. All images have bounding box annotations. Since our classifier works on pixles, we will define our training sets in terms of pixels.

\begin{equation}
\mathcal{T}_S=\left\{(\mathbf{x_1},y_1,z_1),\ldots,(\mathbf{x_m},y_m,z_m)\right\}
\label{eq:synthetic_training_set}
\end{equation}

\begin{equation}
\mathcal{T}_R=\left\{(\mathbf{x_1},z_1),\ldots,(\mathbf{x_k},z_k)\right\}
\label{eq:real_training_set}
\end{equation}

$\mathcal{T}_S$ contains labelled pixles from our synthetic images, where $\mathbf{x_i}$ is a pixel, $y_i$ is one out of 8 body part labels and $z_i$ denotes the spatial bin of the pixel with respect to the bounding box. $\mathcal{T}_R$ contains pixels from our real images, where $\mathbf{x_i}$ and $z_i$ are defined as before.

Furthermore we will denote a node in the random forest by $n$ (see figure \ref{fig:intro:treenode}). The subset of synthetic and real training pixels arriving at node $n$ will be denoted $\mathcal{T}_S(n)$ and $\mathcal{T}_R(n)$, respectively. Finally, we will denote the histogram over body part labels by $\mathbf{h}_l$ and histogram over spatial bins by $\mathbf{h}_s$. For example, $\mathbf{h}_l(\mathcal{T}_S(n))$ is the histogram over body part labels $y_i$ from all synthetic training pixels arriving at node $n$ and $\mathbf{h}_s(\mathcal{T}_R(n))$ is the histogram over spatial bins for all real training pixels arriving at node $n$ (see figures \ref{fig:intro:syn_partdist}, \ref{fig:intro:real_spdist} and \ref{fig:intro:syn_spdist}).


A common approach to training a decision tree is to recursively select the weak classifier that yields the largest drop in entropy of the target label. In this paper we suggest a novel fitness function that, in addition to the label entropy in the synthetic data, also includes the $\chi^2$-distance between the spatial bin histograms of the sets $\mathcal{T}_S(n)$ and $\mathcal{T}_R(n)$. In equation \ref{eq:error} we give the form of the fitness function, $f$, and equation \ref{eq:gain} defines the gain, $I$, of a weak classifier $g$.

\begin{equation} 
\begin{split}
f\left( n \right) =& \alpha \cdot E\left( \mathbf{h}_l\left\{ \mathcal{T}_S(n) \right\} \right) \\
+& \left( 1-\alpha \right) \cdot d_{\chi^2} \left( \mathbf{h}_s\left\{ \mathcal{T}_S(n) \right\}, \mathbf{h}_s\left\{ \mathcal{T}_R(n)\right\} \right)
\label{eq:error}
\end{split}
\end{equation}

\begin{equation} 
IG(g) = f\left(  n \right) - \frac{|\mathcal{T}_S(n_l)|}{|\mathcal{T}_S(n)|}f\left(  n_l \right) - \frac{|\mathcal{T}_S(n_r)|}{|\mathcal{T}_S(n)|}f\left(  n_r \right)
\label{eq:gain}
\end{equation}

Selecting weak classifiers that maximize $IG(g)$ will yield decision trees that produce reasonable output on the real training data, while at the same time minimizing the label entropy on the synthetic test data. We can think of the $d_{\chi^2} \left( \cdot , \cdot \right)$-term of the fitness function as a regularization term, encouraging consistent classification patterns across domains.

\subsection{Weak Classifiers}
\begin{figure}
  	\centering
	\includegraphics[width=0.5\linewidth]{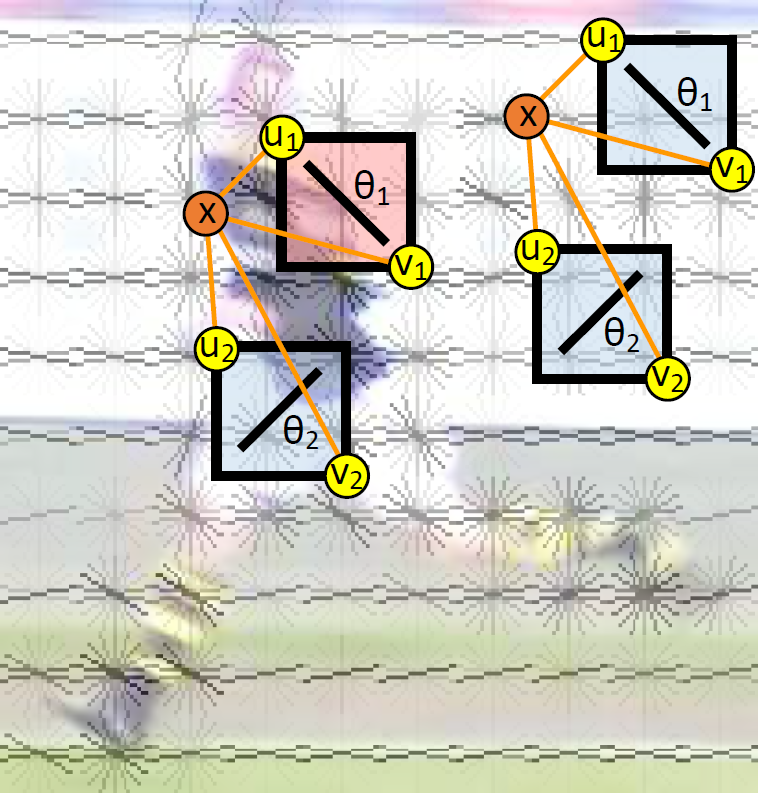}
  	\caption{Illustration of the weak classifier. The classifier is defined by two HOG bins and a threshold. The ratio of the HOG bin responses are compared to the threshold to determine the response of the classifier - see text for details. In the figure two pixels are classified - one of them have a large difference in HOG bin activation because one of the bins align with the contour of the player.}
 	\label{fig:weak_classifier}
\end{figure}

We use weak classifiers based on HOG-like features \cite{Dalal:cvpr:05}. More precisely, the response ratio of two HOG bins are compared to a threshold to determine the response of the classifier, as illustrated in figure \ref{fig:weak_classifier}. A HOG bin is defined by its upper left and lower right corners, $\mathbf{u}$ and $\mathbf{v}$, and a discretized angle $\theta$. The HOG bin response is the weighted, accumulated gradient magnitude of all gradients within the bounding box. If we let $\mathcal{R}(\mathbf{u},\mathbf{v})$ be the set of pixels in the bounding box, the HOG bin response can be defined as in equation \ref{eq:hog_bin_response}.

\begin{equation} 
r(\mathcal{R},\theta) = \sum_{\mathbf{x} \in \mathcal{R}} w(\nabla I_\mathbf{x},\theta) \cdot \| \nabla I_\mathbf{x} \|
\label{eq:hog_bin_response}
\end{equation}

The weight $w(\nabla I_\mathbf{x},\theta) \in \left[0,1\right]$ softly assigns the gradient magnitude to discrete values. The HOG bin respose can be computed efficiently using integral images \cite{Viola:2004:RRF:966432.966458}.

The weak classifier is defined in equation \ref{eq:weak_classifier}, where $\mathbf{x}$ is the classified pixel and $\mathcal{R}_1$, $\theta_1$, $\mathcal{R}_2$, and $\theta_2$ are the parameters of the classifier.

\begin{equation} 
g\left(\mathbf{x};\mathcal{R}_1,\theta_1,\mathcal{R}_2,\theta_2 \right) = sign\left( \frac{ r(\mathcal{R}_1,\theta_1)}{r(\mathcal{R}_2,\theta_2)} - t \right)
\label{eq:weak_classifier}
\end{equation}

\subsection{Implementation details}
\label{sec:impl_details}
There is still a lot of craftsmanship involved in training random forests and training parameter settings can have a large impact on the resulting classifier. In our experiments we used only two decision trees per forest (this is similar to \cite{Shotton:pami:12}, where it was noted that using more trees yielded only a small gain in performance). 

With the weak classifiers used in this paper the training objective function typically leveled off when the decision trees had reached around 12 levels, which is the tree depth we used in our experiments. In initial experiments with other weak classifiers, \eg silhouette-based classifiers, we noticed a need for deeper trees.

We used a breadth first training strategy \cite{MSRDecisionForestBook} to limit memory requirements. We used a two-stage training strategy to train each new level of the tree. The first stage involved making a complete pass through the training data to extract a small uniform sample of labelled and unlabelled training examples at each current leaf node (termed ''frontier node'' in \cite{MSRDecisionForestBook}) of the tree. For this we used reservoir sampling \cite{reservoir_sampling}. We extracted 100 labelled and 100 unlabelled examples for each ''frontier node''. These examples were then used to evaluate a 2000 randomly generated weak classifiers at 60 uniformly sampled thresholds, yielding a total of 120000 unique proposed weak classifiers. The 30 best weak classifiers were selected along with the 10 best thresholds for each weak classifier, yielding a total of 300 selected weak classifiers. In the second stage of trianing, these 300 selected weak classifiers were evaluated on the full training data during a second pass through the training data. The best weak classifier out of the 300 was finally stored in the tree.

\subsection{Converting random forest output to pixel classifications}
\label{sec:pixel_class}
The random forest outputs a probability distribution over the body part labels for every pixel in the image. However, our desired output is a list of pixels for each body part label. To make this conversion we assume that 1) the image covers the whole character and 2) we know the size of the character in the image. The first assumption means that every body part is ''visible'' in the image (\ie a body part may be occluded but the classifier should still be able to locate it in the image). The second assumption means that we know the number, $N$, of pixels covered by each body part. Note that since body parts may be occluding eachother some pixels will have more than one label. Now we can simply select the $N$ pixels with the highest probability for each body part label separately.

\section{Experiments and results}
\label{sec:experiments}
In this section we will evaluate the method by comparing it to the na\"{\i}ve baseline of training only on the synthetic training data and ignoring the real (unlabelled) training data completely. Note that this corresponds to setting $\alpha=1$ in equation \ref{eq:error}. We report results on two experiments. First we use synthetic data for both training and testing and in the second experiment we test on real images \cite{Kazemi2011}. 

The performance measure used for evaluation is the percentage of correctly classified pixels. A pixel classified as, for example, `hip' is counted as correct if it lies within a square with side $L$ centered at the hip location, as illustrated in figure \ref{fig:results_data_c}. This percentage is first computed for each joint label individually and then averaged over joint labels to produce the final performance measure. In all experiments we use $L=0.2\sqrt{A}$, where $A$ is the area of the bounding box (the same setting is used in training).

\subsection{Evaluation on synthetic data}
\label{sec:experiments_synthetic}
\begin{figure}
  	\centering
	\subfigure[]{ \label{fig:results_data_a} \includegraphics[height=0.15\textwidth]{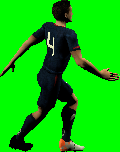} }
	\subfigure[]{ \label{fig:results_data_b} \includegraphics[height=0.15\textwidth]{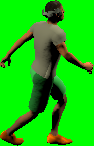} }
	\subfigure[]{ \label{fig:results_data_c} \includegraphics[height=0.15\textwidth]{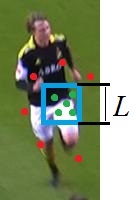} }
  \caption{Characters used to generate datasets for our experiment. a) Character used to generate the labeled training set $\mathcal{T}_S$ for our first experiment. b) Character used to generate unlabeled training set $\mathcal{T}_R$ and test set for our first experiment. c) Evaluation criterion - a predicted class label (e.g. `hip') is counted as correct if it lies within a square with side $L$ centered at the joint location.}
  \label{fig:results_data}
\end{figure}

For this evaluation we employed two synthetic football characters with different clothing and body shape, as shown in figures \ref{fig:results_data_a} and \ref{fig:results_data_b}. We used the first character to generate a set of images with body part labels and bounding box annotations; pixels from those images contstiute  $\mathcal{T}_S$ in this experiment. We used the second character to generate a set of images with only bounding box annotations; pixels from those images constitute $\mathcal{T}_R$ in this experiment. Both image set were generated from the same motion capture sequence, containing a ''walking'' motion. The motion capture sequence contained 112 poses and we used a single viewpoint (from behind), making a total of 112 images in each training set. Finally we used the second character to generate new test images, with poses taken from a different motion capture sequence also depicting ''walking''. This motion capture sequence contained 77 poses and, still rendering images from a single viewpoint, the test set contained 77 images.

\begin{figure}
  	\centering
	\subfigure[$\alpha=0.2$]{ 
		\label{fig:results_synth_train_a}
		\includegraphics[width=0.46\linewidth]{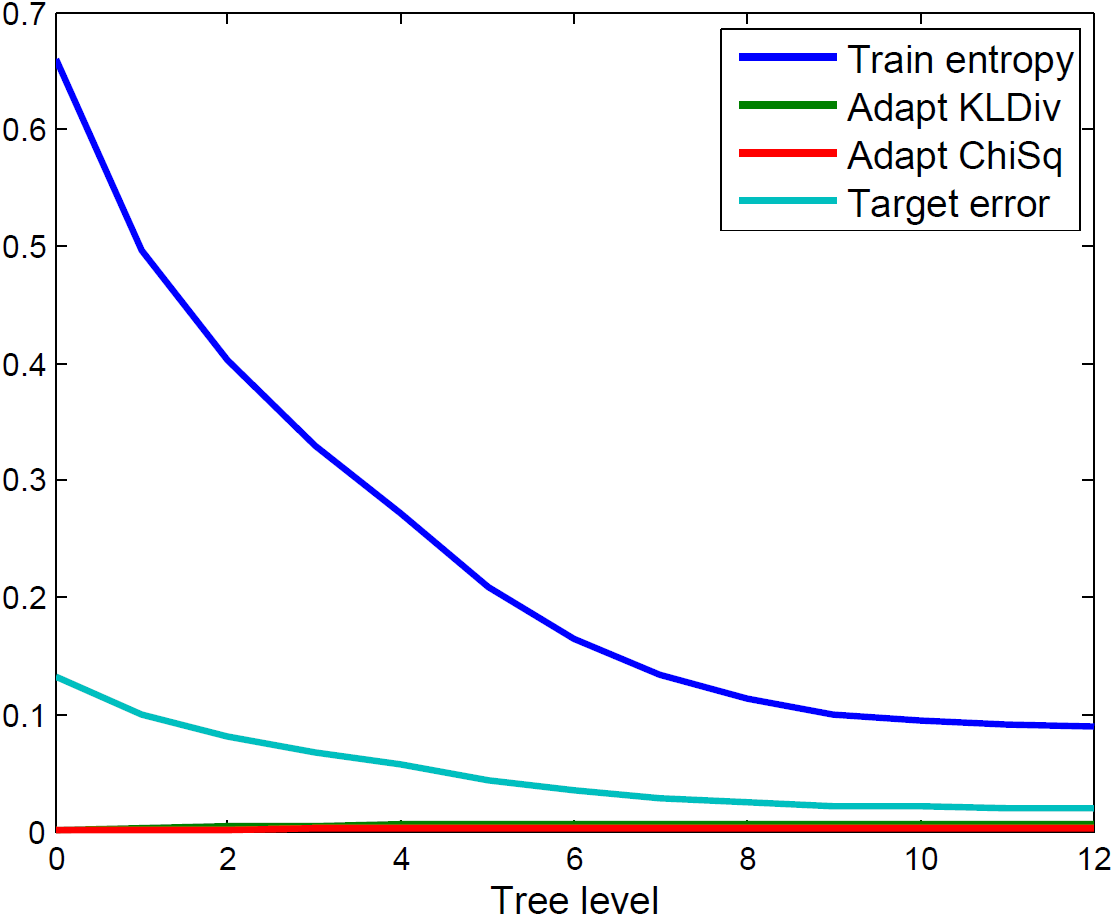} 
	}
	\subfigure[$\alpha=1$]{ 
		\label{fig:results_synth_train_b}
		\includegraphics[width=0.46\linewidth]{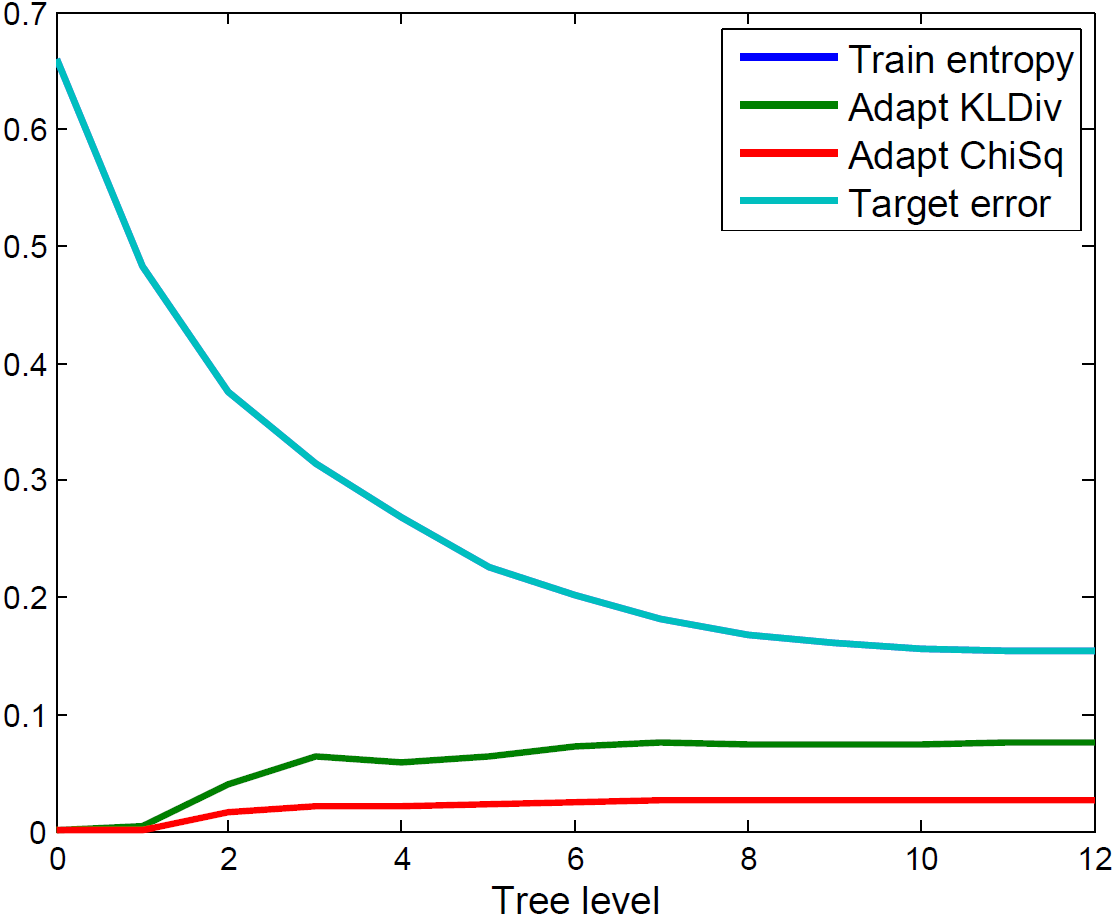} 
	}
  	\caption{Various error measures plotted vs. tree level for $\alpha=0.2$ (a) and $\alpha=1$ (b). We observe that with $\alpha=0.2$ the KL divergence and the $\chi^2$-distance between the sets $\mathcal{T}_S$ and $\mathcal{T}_R$ is kept low and surprisingly we also reach a slightly lower leaf entropy on $\mathcal{T}_S$.}
	\label{fig:results_synth_train}
\end{figure}

We begin by analyzing the results of training the random forest with different values for $\alpha$, see figure \ref{fig:results_synth_train}. Figure \ref{fig:results_synth_train_a} shows that when $\alpha$ is set to a low value ($0.2$) the $\chi^2$-distance and the KL divergence between the spatial distribution of pixels from the two sets $\mathcal{T}_S$ and $\mathcal{T}_R$ are kept low while the entropy of body part label of pixels from the set $\mathcal{T}_S$ is decreased. In figure \ref{fig:results_synth_train_b} we show the same plots for the na\"{\i}ve baseline $\alpha=1$. Note that the target error and the train entropy is the same in this case so the turcose curve is covering the blue one. We see that the spatial distributions diverge as training progresses and interestingly the final train entropy is actually higher in this case than with $\alpha=0.2$. This is a bit unexpected since we are now focusing all our efforts on minimizing the train entropy. One explanation could be that the $d_{\chi^2}(\cdot,\cdot)$-term helps the greedy random forest learner avoid local minima.

\begin{figure}
  	\centering
	\subfigure[Pixels from $\mathcal{T}_S$, $\alpha=0.2$]{ 
 	 	\label{fig:results_synth_training_examples_a}
		\includegraphics[width=0.2\linewidth]{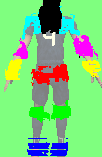}
	}
	\subfigure[Pixels from $\mathcal{T}_S$, $\alpha=1$]{ 
  		\label{fig:results_synth_training_examples_b}
		\includegraphics[width=0.2\linewidth]{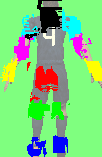}
	}
	\subfigure[Pixels from $\mathcal{T}_R$, $\alpha=0.2$]{ 
  		\label{fig:results_synth_training_examples_c}
		\includegraphics[width=0.2\linewidth]{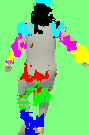}
	}
	\subfigure[Pixels from $\mathcal{T}_R$, $\alpha=1$]{ 
  		\label{fig:results_synth_training_examples_d}
		\includegraphics[width=0.2\linewidth]{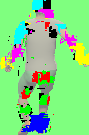}
	}
  \caption{Qualitative illustration of the performance on the training sets $\mathcal{T}_S$ and $\mathcal{T}_R$ for different values of $\alpha$. The images show classification results on a) an image from the labelled training set $\mathcal{T}_S$ with $\alpha=0.2$ and b) with $\alpha=1$ and c) an image from the unlabelled training set $\mathcal{T}_R$ with $\alpha=0.2$ and d) with $\alpha=1$. Pixel classifications were extracted according to section \ref{sec:pixel_class}.}
  \label{fig:results_synth_training_examples}
\end{figure}

In figure \ref{fig:results_synth_training_examples} we show some example classifications on the training sets $\mathcal{T}_S$ and $\mathcal{T}_R$ for random forests trained with $\alpha=0.2$ and $\alpha=1$. Comparing \ref{fig:results_synth_training_examples_a} and \ref{fig:results_synth_training_examples_b} we get a qualitative confirmation that using a lower value for alpha can actually improve the classification performace even on the training set $\mathcal{T}_S$. Figures \ref{fig:results_synth_training_examples_c} and \ref{fig:results_synth_training_examples_d} show qualitative classification results on the unlabelled training set $\mathcal{T}_R$. 

Moving on to analyze the performance on test data, we give the percent of correctly classified pixels in the test set for different values of $\alpha$ in table \ref{tab:results_synth}. The first row of the table gives the value of $\alpha$ used during training. The second row gives the weighted average leaf-node entropy of body part labels for the labelled training set $\mathcal{T}_S$. Again we can see the trend of lower leaf entropy for lower values of $\alpha$. The third row gives the percent of correctly classified pixels on the test set. The best performance, 77\%, was achieved for $\alpha=0.2$, compared to only 43\% for the baseline $\alpha=1$. We also estimated a location prior from the labelled training set and when that is used to modulate the output of the random forest classifier, we achieve the performance $p^{\prime}$ given in the last row of the table. Using the prior alone gives a performance of 79\%. Since the random forest classifier only uses local image appearance to classify a pixel, the location prior provides complementary information and is a simple way of boosting performance.

\begin{table}
\centering
\tabcolsep=0.11cm
\begin{tabular}{ |l|l|l|l|l|l|l|l|l|l|l|l|l| }
\hline
$\alpha$&		0.1& 		\textbf{0.2}& 		0.3& 		0.4& 		0.5& 		0.6&		0.7&		0.8&		0.9&		1.0\\
$e_{leaf}$&	0.09&	\textbf{0.09}&		0.09&	0.11&	0.14&	0.11&	0.13&	0.22&	0.18&	0.15\\
$p$ & 		72& 		\textbf{77}& 		68& 		68& 		69&		65&		61&		62&		59&		43 \\
$p^{\prime}$ &	84&		\textbf{85}&		84&		84&		83&		83&		82&		\textbf{89}&		78&		74 \\
\hline
\end{tabular}
\caption{Weighted average leaf-node entropy $e_{leaf}$ (the root node entropy is 0.66) and percent correctly classified pixels $p$ on the \emph{test set} for different values of the $\alpha$ parameter. $p^{\prime}$ is the percent correctly classified pixels when we modulate the random forest output by joint location priors estimated from the labelled training set $\mathcal{T}_S$.}
\label{tab:results_synth}
\end{table}

\begin{figure}
  	\centering
	\subfigure[Input image]{ 
  		\label{fig:results_synth_ex_a}
		\includegraphics[height=0.3\linewidth]{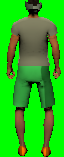}
		\includegraphics[height=0.3\linewidth]{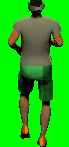}
		\includegraphics[height=0.3\linewidth]{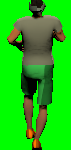} 
		\includegraphics[height=0.3\linewidth]{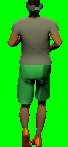} 
		\includegraphics[height=0.3\linewidth]{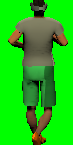} 
	} \\

	\subfigure[Classifier output, $\alpha=1$]{ 
  		\label{fig:results_synth_ex_c}
		\includegraphics[height=0.3\linewidth]{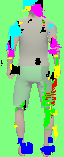}
		\includegraphics[height=0.3\linewidth]{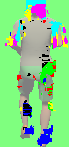}
		\includegraphics[height=0.3\linewidth]{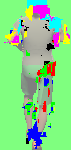} 
		\includegraphics[height=0.3\linewidth]{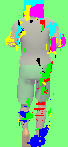} 
		\includegraphics[height=0.3\linewidth]{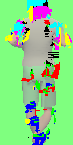} 
	} \\

	\subfigure[Classifier output, $\alpha=0.2$]{ 
  		\label{fig:results_synth_ex_b}
		\includegraphics[height=0.3\linewidth]{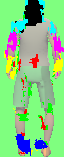}
		\includegraphics[height=0.3\linewidth]{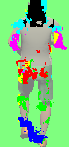}
		\includegraphics[height=0.3\linewidth]{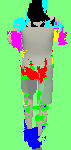} 
		\includegraphics[height=0.3\linewidth]{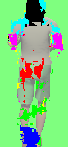} 
		\includegraphics[height=0.3\linewidth]{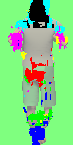} 
	} \\

	\subfigure[Estimated joint locations, $\alpha=0.2$]{ 
  		\label{fig:results_synth_ex_d}
		\includegraphics[height=0.3\linewidth]{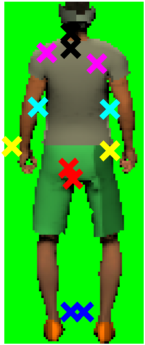}
		\includegraphics[height=0.3\linewidth]{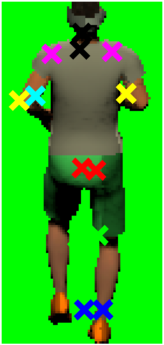}
		\includegraphics[height=0.3\linewidth]{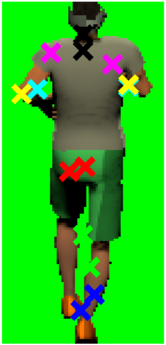} 
		\includegraphics[height=0.3\linewidth]{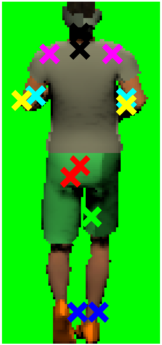} 
		\includegraphics[height=0.3\linewidth]{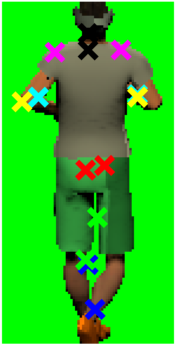} 
	} \\
  \caption{Comparison of joint classification output on a few example images from the test set (a), for the baseline classifier $alpha=1$ (b) and for the classifier trained with $alpha=0.2$ (c). Pixel classifications were extracted according to section \ref{sec:pixel_class}. In (d) we show estimated joint locations exploiting the additional information from the location priors learned from the labelled training set $\mathcal{T}_S$.}
  \label{fig:results_synth_ex}
\end{figure}

Finally, we show a few example classifications on images from the test set in figure \ref{fig:results_synth_ex}. Qualitatively we can see that the poor performance of the classifier trained with $\alpha=1$ (figure \ref{fig:results_synth_ex_c}) renders it almost useless in practice, while the classifier trained with $\alpha=0.2$ (figure \ref{fig:results_synth_ex_b}) can be used quite effectively for joint localization. In figure \ref{fig:results_synth_ex_d} we show estimated joint locations, using the random forest likelihood modulated by the location prior.

\subsection{Evaluation on real data}
\label{sec:experiments_real}
This experiment is very similar to the previous one, with the exception that we used real images to populate the unlabeled training set $\mathcal{T}_R$ and we also used real images for test. Labelled training pixels were still extracted from synthetic images. We used the same character as in the previous experiment (figure \ref{fig:results_data_a}), but a different motion capture sequence. The motion capture used for this experiment contained 275 poses (depicting ''walking'' and ''running'') and each pose was rendered from 16 different viewpoints, yielding a total of 4400 labelled, synthetic images.

The unlabelled training data $\mathcal{T}_R$ was taken from the KTH multiview football dataset \cite{Kazemi2011}, containing 771 annotated and cropped images of football players in action. As suggested in \cite{Kazemi2011}, we use the first 180 images in the dataset for training. Note, however, that we do \emph{not} use the labels provided with the data. The remaining 591 images were used for testing.

\begin{figure}
  	\centering
	\subfigure[$\alpha=0.2$]{ \includegraphics[width=0.46\linewidth]{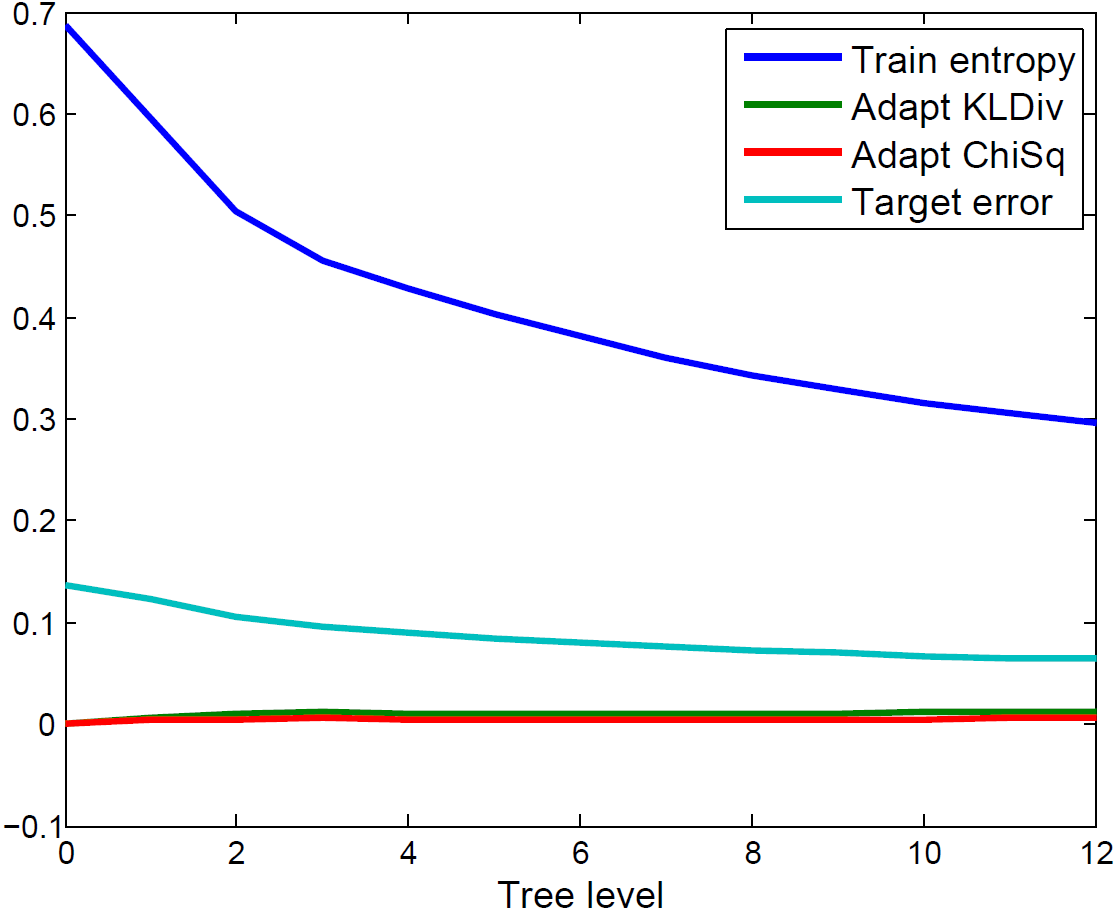} }
	\subfigure[$\alpha=1$]{ \includegraphics[width=0.46\linewidth]{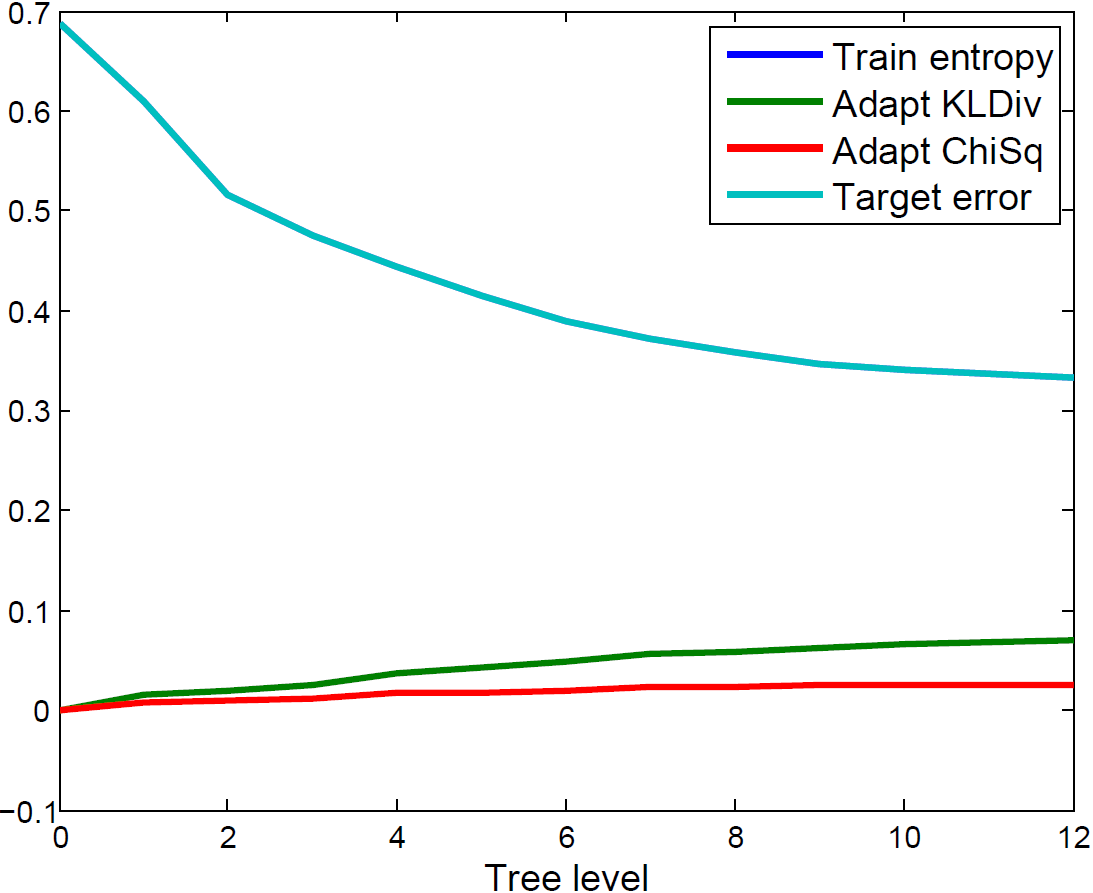} }
  \caption{Various error measures plotted vs. tree level for $\alpha=0.2$ (a) and $\alpha=1$ (b). We observe that with $\alpha=0.2$ the KL divergence and the $\chi^2$-distance between the sets $\mathcal{T}_S$ and $\mathcal{T}_R$ is kept low and surprisingly, as in the previous experiment, we again reach a slightly lower leaf entropy on $\mathcal{T}_S$.}
  \label{fig:results_real_train}
\end{figure}

As in the previous experiment, we begin by analyzing the results of training the random forest with different values for $\alpha$, see figure \ref{fig:results_real_train} and compare to figure \ref{fig:results_synth_train}. As before we observe that when $\alpha$ is set to a low value ($0.2$) the $\chi^2$-distance and the KL divergence are kept low while with $\alpha=1$  the spatial distributions diverge. Again the final train entropy is lower with $\alpha=0.2$ than with $\alpha=1$. 

\begin{figure}
  	\centering
	\subfigure[Pixels from $\mathcal{T}_S$, $\alpha=0.2$]{ 
  		\label{fig:results_real_training_examples_a}
		\includegraphics[width=0.2\linewidth]{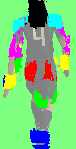}
	}
	\subfigure[Pixels from $\mathcal{T}_S$, $\alpha=1$]{ 
  		\label{fig:results_real_training_examples_b}
		\includegraphics[width=0.2\linewidth]{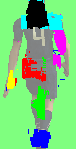}
	}
	\subfigure[Pixels from $\mathcal{T}_R$, $\alpha=0.2$]{ 
  		\label{fig:results_real_training_examples_c}
		\includegraphics[width=0.2\linewidth]{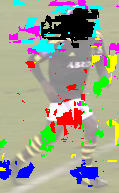}
	}
	\subfigure[Pixels from $\mathcal{T}_R$, $\alpha=1$]{ 
  		\label{fig:results_real_training_examples_d}
		\includegraphics[width=0.2\linewidth]{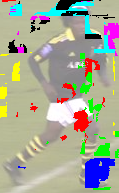}
	}
  \caption{Qualitative illustration of the performance on the training sets $\mathcal{T}_S$ and $\mathcal{T}_R$ for different values of $\alpha$. The images show classification results on a) an image from the labelled training set $\mathcal{T}_S$ with $\alpha=0.2$ and b) with $\alpha=1$ and c) an image from the unlabelled training set $\mathcal{T}_R$ with $\alpha=0.2$ and d) with $\alpha=1$. Pixel classifications were extracted according to section \ref{sec:pixel_class}}
  \label{fig:results_real_training_examples}
\end{figure}

In figure \ref{fig:results_real_training_examples} we again observe that setting $\alpha=0.2$ yields a qualitative improvement in classification performance on both training sets  $\mathcal{T}_S$ and $\mathcal{T}_R$ compared to $\alpha=1$. Figures \ref{fig:results_real_training_examples_a} and \ref{fig:results_real_training_examples_b} show classifications on labelled training images while figures \ref{fig:results_real_training_examples_c} and \ref{fig:results_real_training_examples_d} show classifications on unlabelled training examples.

Regarding performance on test data, the best result, $p=$33\%, was achieved for $\alpha=0.2$, compared to $p=$21\% for the baseline $\alpha=1$ (table \ref{tab:results_real}). This confirms the results from the previous section, albeit with overall lower performance. The difference between the source and target domains is in this experiment much larger than in the previous one. In addition, this experiment presented a much more varied dataset, containing many more poses and viewpoints than the dataset used in the previous experiment. Again the last row of the table shows that a performance boost is possible by exploiting the location prior. The performance using the prior alone was in this experiment 45\%.

\begin{table}
\centering
\tabcolsep=0.11cm
\begin{tabular}{ |l|l|l|l|l|l|l|l|l|l|l|l|l| }
\hline
$\alpha$&		0.1& 		\textbf{0.2}& 		0.3& 		0.4& 		0.5& 		0.6&		0.7&		0.8&		0.9&		1.0\\
$e_{leaf}$&	0.32&	\textbf{0.30}&		0.30&	0.31&	0.30&	0.31&	0.33&	0.31&	0.34&	0.33\\
$p$ & 		32& 		\textbf{33}& 		31& 		30& 		27&		27&		25&		26&		23&		21 \\
$p^{\prime}$ & 46&		\textbf{48}&		46&		47&		44&		46&		42&		43&		45&		43 \\
\hline
\end{tabular}
\caption{Weighted average leaf-node entropy $e_{leaf}$ (the root node entropy is 0.69) and percent correctly classified pixels $p$ on the \emph{test set} for different values of the $\alpha$ parameter. $p^{\prime}$ is the percent correctly classified pixels when we also estimate location priors from the annotated training set $\mathcal{T}_S$.}
\label{tab:results_real}
\end{table}

\begin{figure}
  	\centering
	\subfigure[Input image]{ 
		\includegraphics[height=0.3\linewidth]{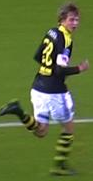}
		\includegraphics[height=0.3\linewidth]{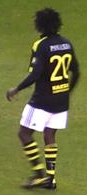}
		\includegraphics[height=0.3\linewidth]{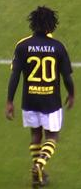} 
		\includegraphics[height=0.3\linewidth]{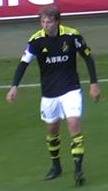} 
		\includegraphics[height=0.3\linewidth]{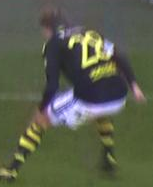} 
	} \\

	\subfigure[Classifier output, $\alpha=1$]{ 
		\includegraphics[height=0.3\linewidth]{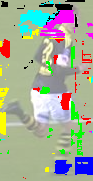}
		\includegraphics[height=0.3\linewidth]{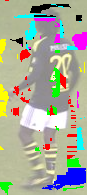}
		\includegraphics[height=0.3\linewidth]{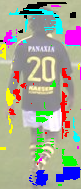} 
		\includegraphics[height=0.3\linewidth]{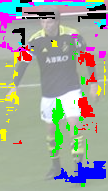} 
		\includegraphics[height=0.3\linewidth]{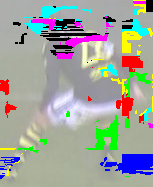} 
	} \\

	\subfigure[Classifier output, $\alpha=0.2$]{ 
		\includegraphics[height=0.3\linewidth]{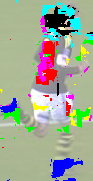}
		\includegraphics[height=0.3\linewidth]{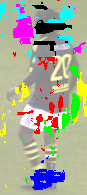}
		\includegraphics[height=0.3\linewidth]{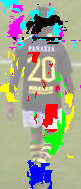} 
		\includegraphics[height=0.3\linewidth]{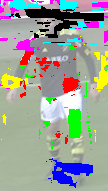} 
		\includegraphics[height=0.3\linewidth]{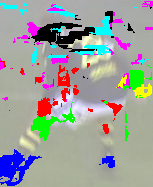} 
	} \\

	\subfigure[Estimated joint locations, $\alpha=0.2$]{ 
		\includegraphics[height=0.3\linewidth]{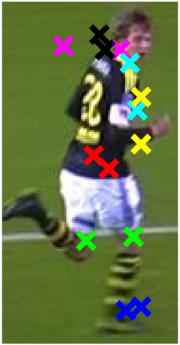}
		\includegraphics[height=0.3\linewidth]{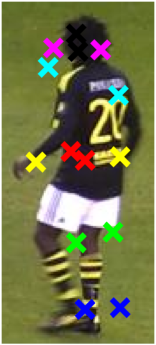}
		\includegraphics[height=0.3\linewidth]{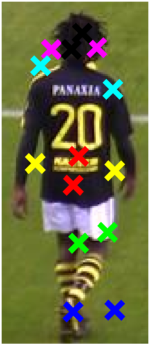} 
		\includegraphics[height=0.3\linewidth]{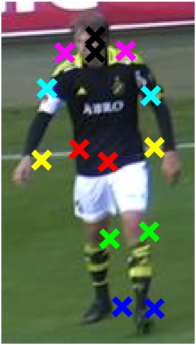} 
		\includegraphics[height=0.3\linewidth]{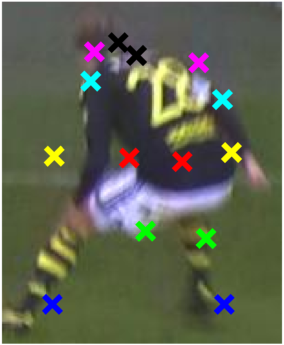} 
	}
  \caption{Comparison of joint classification output on a few example images from the test set (a), for the baseline classifier $alpha=1$ (b) and for the classifier trained with $alpha=0.2$ (c). Pixel classifications were extracted according to section \ref{sec:pixel_class}. In (d) we show estimated joint locations exploiting the additional information from the location priors learned from the labelled training set $\mathcal{T}_S$.}
  \label{fig:results_real_ex}
\end{figure}

In figure \ref{fig:results_real_ex} show some example classifications on images from the test set. It is apparent that the large domain difference in this case makes it hard for the classifier to correctly locate some of the body parts, especially the ''interior'' parts: `hip', `sholder' and `elbow'. However, the ''extremities'': `head', `foot', `hand' and `knee', are often detected correctly. The baseline $\alpha=1$ typically only succeds in locating the `foot'.  

\section{Acknowledgements}
This work was supported by the FP7 project "Free-viewpoint Immersive Networked Experience" and ChyronHego AB.

\section{Conclusion}
\label{sec:conclusion}
Random forests are a great tool for solving complex classification tasks due to their ability to represent complex decision boundaries. However, they overfit to the training data. Therefore the training data is required to 1) come from the same distribution as the test data and 2) represent all the variation in the test data. We have addressed the second constraint by using computer graphics generated, synthetic training data so that we can explicitly control the diversity in the training data. The first constraint is in practice very expensive and often impossible to satisfy, motivating our efforts to use weakly labeled training data from the target domain to encourage the learner to generalize to the target domain. We have shown, on both synthetic and real target domains, that modifying the objective function for random forest training to encourage consistent spatial node activation patterns across domains siginficantly improves classification performance on the test data. Interestingly, it also yielded a slightly lower leaf entropy in body part label on the training set. This was unexpected, since we are actually putting less emphasis on minimizing the entropy. One possible explanation is that the modified objective function is less likely to fall into local minima.

While we have proposed and evaluated this method for pose estimation from RGB images, it is possible to generalize it to other application domains as well. Where we used the distance between the spatial distributions of pixels from the labelled and unlabelled training sets ($\mathcal{T}_S$ and $\mathcal{T}_R$) , one could use the distribution of some other quatity that is correlated to the target label. This is an interesting direction for future work.

\bibliographystyle{IEEEtranS}
\small{
\bibliography{bibl}
}

\end{document}